\documentclass[conference]{IEEEtran}
\IEEEoverridecommandlockouts
\usepackage{cite}
\usepackage{amsmath,amssymb,amsfonts}
\usepackage{algorithmic}
\usepackage{graphicx}
\usepackage{textcomp}
\usepackage{xcolor}
\usepackage{array}
\usepackage{float}
\usepackage{url}

\def\BibTeX{{\rm B\kern-.05em{\sc i\kern-.025em b}\kern-.08em
    T\kern-.1667em\lower.7ex\hbox{E}\kern-.125emX}}
\begin{document}

\title{Anti-Jamming based on Null-Steering Antennas and Intelligent UAV Swarm Behavior\\
\thanks{This work is funded by national funds through FCT — Fundação para a Ciência e a Tecnologia, I.P., under projects/supports UID/6486/2025 (https://doi.org/10.54499/UID/06486/2025) and UID/PRR/6486/2025 (https://doi.org/10.54499/UID/PRR/06486/2025).}
}

\author{\IEEEauthorblockN{1\textsuperscript{st} Miguel Lourenço}
\IEEEauthorblockA{\textit{Instituto Superior Técnico, University of Lisbon}\\
Lisbon, Portugal \\
miguel.c.lourenco@tecnico.ulisboa.pt}
\and
\IEEEauthorblockN{2\textsuperscript{nd} António Grilo}
\IEEEauthorblockA{\textit{INESC INOV} \\
\textit{Instituto Superior Tecnico, University of Lisbon}\\
Lisbon, Portugal \\
antonio.grilo@inov.pt}

}

\maketitle

\begin{abstract}
Unmanned Aerial Vehicle (UAV) swarms represent a key advancement in autonomous systems, enabling coordinated missions through inter-UAV communication. However, their reliance on wireless links makes them vulnerable to jamming, which can disrupt coordination and mission success. This work investigates whether a UAV swarm can effectively overcome jamming while maintaining communication and mission efficiency.

To address this, a unified optimization framework combining Genetic Algorithms (GA), Supervised Learning (SL), and Reinforcement Learning (RL) is proposed. The mission model, structured into epochs and timeslots, allows dynamic path planning, antenna orientation, and swarm formation while progressively enforcing collision rules. Null-steering antennas enhance resilience by directing antenna nulls toward interference sources.

Results show that the GA achieved stable, collision-free trajectories but with high computational cost. SL models replicated GA-based configurations but struggled to generalize under dynamic or constrained settings. RL, trained via Proximal Policy Optimization (PPO), demonstrated adaptability and real-time decision-making with consistent communication and lower computational demand. Additionally, the Adaptive Movement Model generalized UAV motion to arbitrary directions through a rotation-based mechanism, validating the scalability of the proposed system.

Overall, UAV swarms equipped with null-steering antennas and guided by intelligent optimization algorithms effectively mitigate jamming while maintaining communication stability, formation cohesion, and collision safety. The proposed framework establishes a unified, flexible, and reproducible basis for future research on resilient swarm communication systems.
\end{abstract}

\begin{IEEEkeywords}
UAV Swarm, Anti-Jamming, Genetic Algorithm (GA), Supervised Learning (SL), Reinforcement Learning (RL), Path Planning, Antenna Orientation, Null-Steering, Swarm Formation, Communication Resilience.
\end{IEEEkeywords}

\section{Introduction}

\begin{figure}[htbp]
    \centering
    \includegraphics[width=0.45\textwidth]{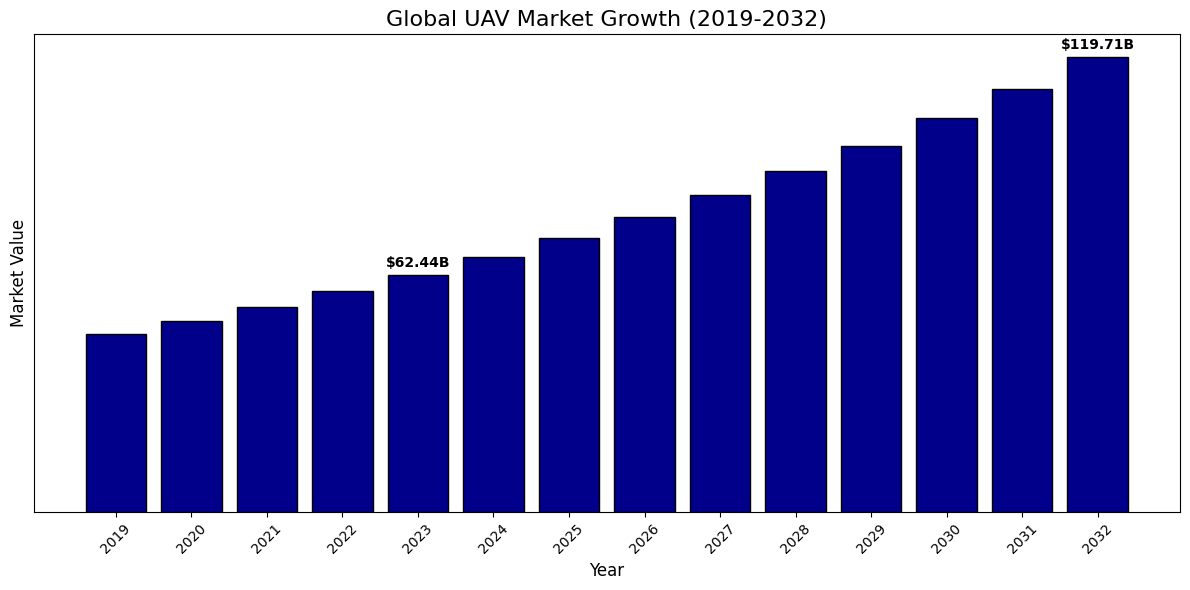}
    \caption{Global UAV Market Growth (2019–2032), adapted from~\cite{b1}.}
    \label{fig:drones_growt}
\end{figure}

The Unmanned Aerial Vehicle (UAV) market has experienced exponential growth over the last decade, underscoring their increasing importance across sectors such as logistics, surveillance, agriculture, and military operations. As illustrated in Fig.~\ref{fig:drones_growt}, the global UAV market value rose from \$62.44 billion in 2023 and is projected to reach \$119.71 billion by 2032~\cite{b1}. This rapid expansion is driven by technological progress, operational efficiency demands, and the unique capability of UAVs to perform critical tasks with precision and reliability. Their adoption has transformed modern operations, making UAVs indispensable assets in today’s world.

Although UAVs represent a modern technological achievement, the challenges they face, particularly communication interference, have deep historical roots. The use of jamming as a strategic tactic dates back to early electronic warfare. During World War II, for instance, the ``Battle of the Beams'' saw Allied forces disrupt German radio signals to hinder bombing missions, revealing the importance of communication dominance in warfare. Initially deployed for reconnaissance in mid-20th century operations, UAVs have since evolved into autonomous, multi-purpose systems. Contemporary conflicts, such as the Russia-Ukraine war, highlight their dual role in surveillance and offense, while also exposing their susceptibility to Radio Frequency (RF) jamming and electronic interference, leading to communication losses, mission failures, and strategic vulnerabilities.

As the UAV market expands and their role in critical operations grows, the need to secure communication systems against interference becomes increasingly urgent. Adversaries can exploit these vulnerabilities to neutralize UAV advantages, emphasizing the necessity of resilient anti-jamming strategies to ensure mission success and operational safety. This study is motivated by the growing intersection between the technological evolution of UAVs and the rising complexity of electromagnetic threats.

\subsection{Motivation}

UAV swarms represent a major step in autonomous systems, relying on inter-UAV communication to execute coordinated missions efficiently. Robust and uninterrupted communication networks are essential for synchronization, information exchange, and real-time adaptation to dynamic environments. However, this dependence on communication makes UAV swarms especially vulnerable to jamming or other deliberate interferences that can disrupt links and compromise coordination.

\begin{figure}[htbp]
    \centering
    \includegraphics[width=0.45\textwidth]{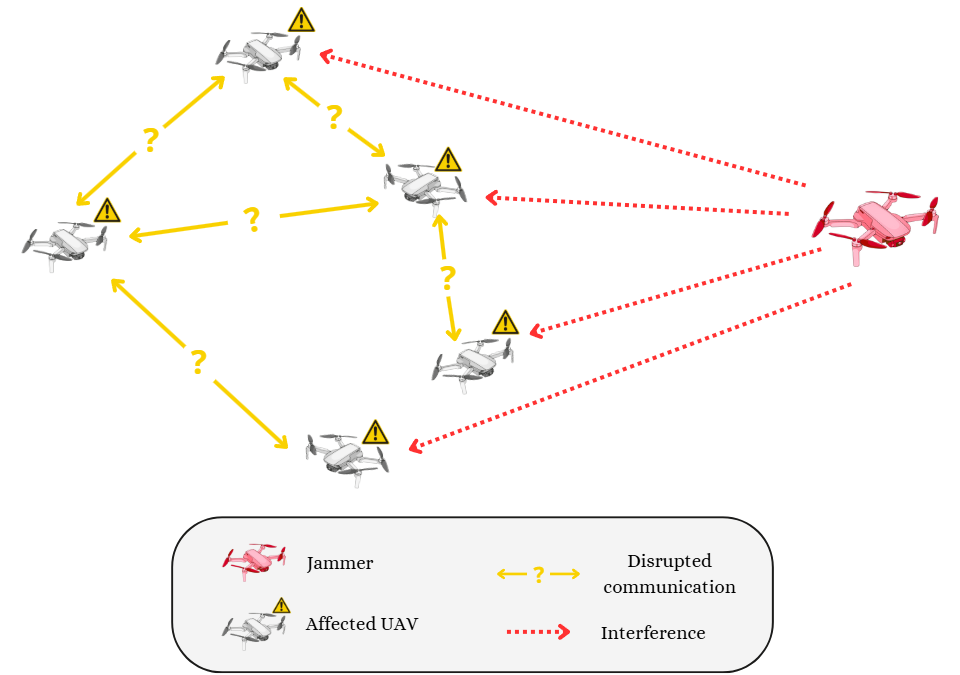}
    \caption{Impact of jamming on inter-UAV communication.}
    \label{fig:motivation}
\end{figure}

As shown in Fig.~\ref{fig:motivation}, the presence of a jammer can disrupt communication links within the swarm, represented by the yellow arrows with question marks. This interference hinders coordination, potentially leading to disorganized behavior, mission failure, or even collisions. The consequences are more severe in swarms than in single-UAV operations, since a single disrupted link can cascade into network-wide communication loss. This vulnerability underscores the need for dedicated anti-jamming techniques tailored for UAV swarms, ensuring reliable operation even under deliberate adversarial interference.

\subsection{Objectives}

Based on the motivation outlined above, this research seeks to answer a central question: 
\textit{Can a swarm of UAVs effectively overcome jamming while maintaining communication and achieving mission success?}

To address this question, the main objective of this work is to develop a unified anti-jamming framework for UAV swarms that integrates null-steering antennas with swarm intelligence techniques, leveraging Genetic Algorithms (GA), Supervised Learning (SL), and Reinforcement Learning (RL). The goal is not only to provide a new perspective on swarm-based communication resilience but also to improve the efficiency and adaptability of UAV coordination under adversarial interference.

Specifically, this study focuses on optimizing key aspects of swarm operation, including antenna directionality, swarm formation, and path-planning, while considering dynamic mission progression across multiple epochs and timeslots. Unlike previous approaches that address these parameters independently or under static assumptions, the proposed method integrates them into a single adaptive framework. Through comparative analysis of GA, SL, and RL under identical conditions, this work aims to evaluate their relative effectiveness in enhancing communication robustness, collision avoidance, and mission success under jamming conditions.

\section{Background and Related Work}
\label{section:background}

\subsection{Metaheuristics}
\label{subsec:meta}

Metaheuristics are high-level optimization strategies that efficiently explore large search spaces to find near-optimal solutions. Among them, Genetic Algorithms (GA) are particularly effective for complex problems due to their balance between exploration and exploitation \cite{Cicirello2024}.

\subsubsection{Genetic Algorithms (GA)}
\label{subsec:ga}

The Genetic Algorithm (GA) \cite{Katoch2021} is an evolutionary optimization method inspired by natural selection. It evolves a population of candidate solutions through \textit{selection}, \textit{crossover}, and \textit{mutation}, guided by a fitness function that evaluates performance.

In this research, GA serves as the baseline optimizer to determine optimal UAV positions and antenna orientations under jamming. It also generates reference datasets later used to train supervised and reinforcement learning models. Although effective, its high computational cost motivates the use of learning-based methods for faster adaptation in subsequent stages.

\subsection{Supervised Learning (SL)}
\label{subsec:sl}

Supervised learning (SL) is a data-driven approach where models learn to map inputs to outputs using labeled datasets. In this work, SL algorithms are trained on GA-generated data to predict UAV positions and antenna orientations that maximize communication capacity under jamming conditions.

\subsubsection{K-Nearest Neighbors (KNN)}
\label{subsec:knn}

KNN \cite{Suyal2022} estimates a target value based on the \(k\) nearest samples in the training set, using distance-weighted averaging.  
It performs local interpolation of GA-generated solutions, making it suitable for approximating discrete optimization results in continuous spatial domains.

\subsubsection{Random Forest (RF)}
\label{subsec:rf}

RF \cite{Salman2024} is an ensemble of decision trees trained on random data subsets.  
It captures non-linear dependencies and generalizes well to unseen configurations, effectively modeling relationships between UAV and jammer positions.

\subsubsection{XGBoost}
\label{subsec:xgb}

XGBoost \cite{xgboost} is a gradient boosting framework that sequentially combines weak learners to minimize prediction error.  
It serves as a benchmark to evaluate whether advanced boosting techniques can outperform simpler ensemble or interpolation methods in predicting optimal UAV configurations.

\subsection{Reinforcement Learning (RL)}
\label{subsec:rl}

Reinforcement Learning (RL) enables an agent to learn optimal sequential actions through interaction with its environment, receiving rewards that guide behavior improvement \cite{SuttonRL}.  
Formulated as a Markov Decision Process (MDP), RL seeks an optimal policy \(\pi^*(a|s)\) that maximizes the expected cumulative reward by balancing exploration and exploitation.

Among various RL algorithms, Proximal Policy Optimization (PPO) \cite{schulman2017proximalpolicyoptimizationalgorithms} is a stable and efficient policy-gradient algorithm for continuous control tasks.  
It limits abrupt policy updates through a clipping mechanism in its objective function, ensuring smooth convergence and robust learning.  
Due to its balance of simplicity and reliability, PPO is widely used in UAV coordination and adaptive control problems.

\section{Related Work}

From the reviewed literature, it becomes clear that significant progress has been made in UAV swarm communication, path planning, and anti-jamming strategies. However, most existing works address these challenges in isolation - focusing either on path optimization, jamming mitigation, or swarm coordination - rather than on their combined and interdependent nature.

While studies such as \cite{bhunia2018distributed} and \cite{Peng2019} demonstrate the effectiveness of null-steering antennas and multi-parameter adaptation, they often rely on static or simplified environments that do not reflect realistic mission progression or dynamic swarm behavior. Similarly, GA-based methods (\cite{Pehlivanoglu2021}, \cite{wanglisong} ) show strong optimization capabilities but are limited by computational cost and weak generalization to unseen conditions. Reinforcement learning approaches (\cite{Wang2023}, \cite{Ye2023} ) introduce adaptability but are typically evaluated in scenarios that exclude collisions or assume idealized communication models.

Moreover, few works explicitly analyze the impact of collisions  on communication performance, despite collisions being a natural and critical constraint in dense UAV swarms. The absence of comparative studies across paradigms (GA, SL, and RL) under consistent experimental conditions further limits the understanding of their relative strengths and weaknesses.

\section{Methodology}

This section presents the proposed methodology for enhancing UAV swarm resilience against jamming interference. It introduces the system model, defines the problem formulation, and outlines the algorithmic framework integrating Genetic Algorithms (GA), Supervised Learning (SL), and Reinforcement Learning (RL). Finally, the evaluation metrics and test scenarios are described.

\subsection{System Model}

Modern UAV swarms must operate in contested environments where radio interference and jamming can severely degrade communication. The proposed system combines adaptive null-steering antennas with swarm coordination and path-planning strategies to mitigate such interference. 

The mission trajectory is divided into multiple \textit{decision epochs}, each consisting of $T+1$ timeslots ($t_0$ to $t_T$). At each epoch, UAVs evaluate their performance via the objective function defined in Section~\ref{subsec:problem}, determining the best parameters for the subsequent epoch. This iterative process continues until mission completion.

\begin{figure}[htbp]
    \centering
    \includegraphics[width=0.45\textwidth]{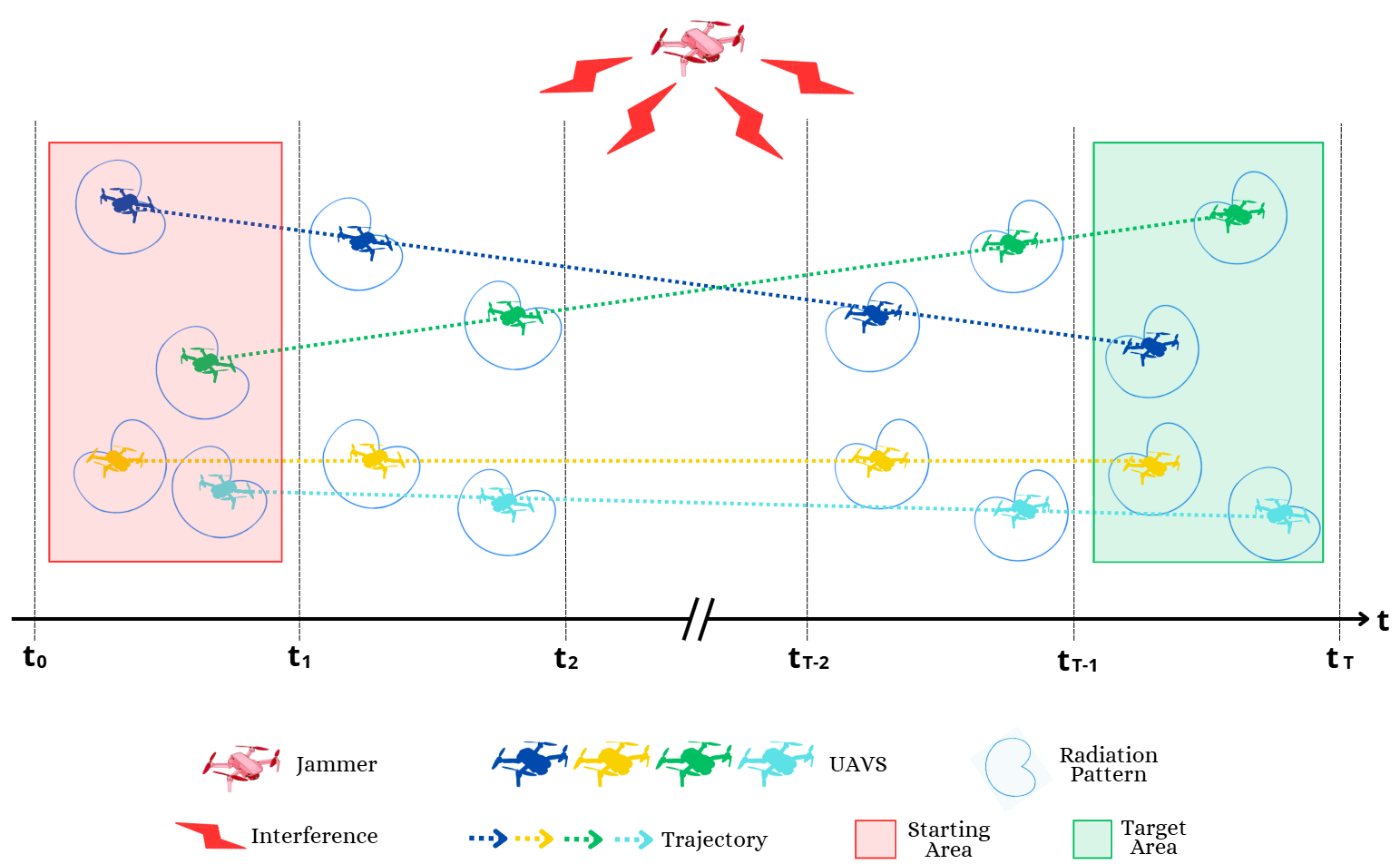}
    \caption{Schematic representation of UAV swarm decision epoch.}
    \label{fig:jammingcomm}
\end{figure}

Each UAV operates within a moving rectangular area, maintaining constant $y$-limits while $x$-bounds progress forward, forming a dynamic operational corridor. UAVs are equipped with null-steering antennas that direct radiation nulls toward interference sources while collaboratively adapting positions to sustain connectivity (Fig.~\ref{fig:jammingcomm}).

\begin{figure}[htbp]
    \centering
    \includegraphics[width=0.35\textwidth]{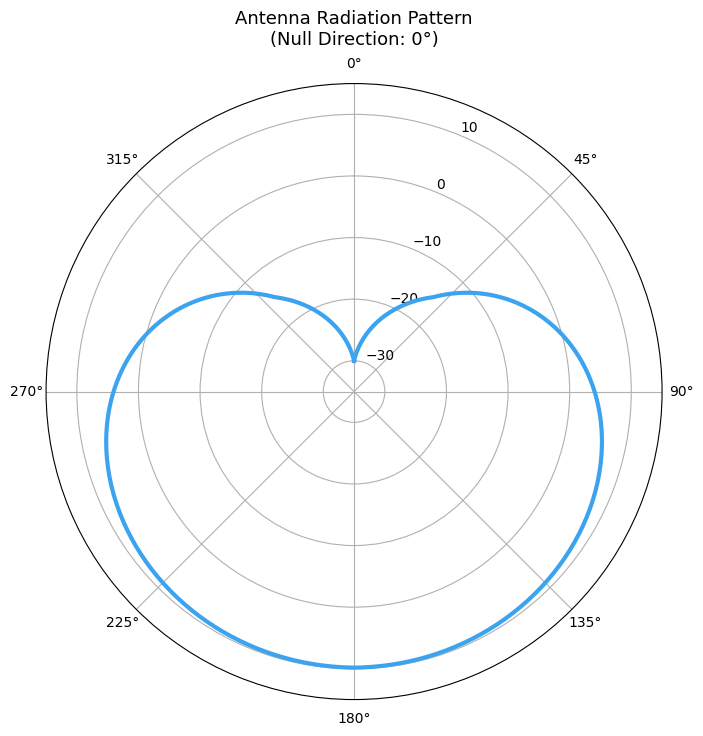}
    \caption{Antenna radiation pattern with null steering.}
    \label{fig:radpat}
\end{figure}

Jammers are modeled with directional antennas to maximize interference, while UAVs employ null-steering antennas to suppress it. The antenna radiation pattern, implemented in a script, adjusts gain according to the null direction, as shown in Fig.~\ref{fig:radpat}.

This model enables realistic optimization of communication under jamming while maintaining flexibility in multi-node links.

\subsection{Problem Formulation}
\label{subsec:problem}

\begin{figure}[htbp]
    \centering
    \includegraphics[width=0.45\textwidth]{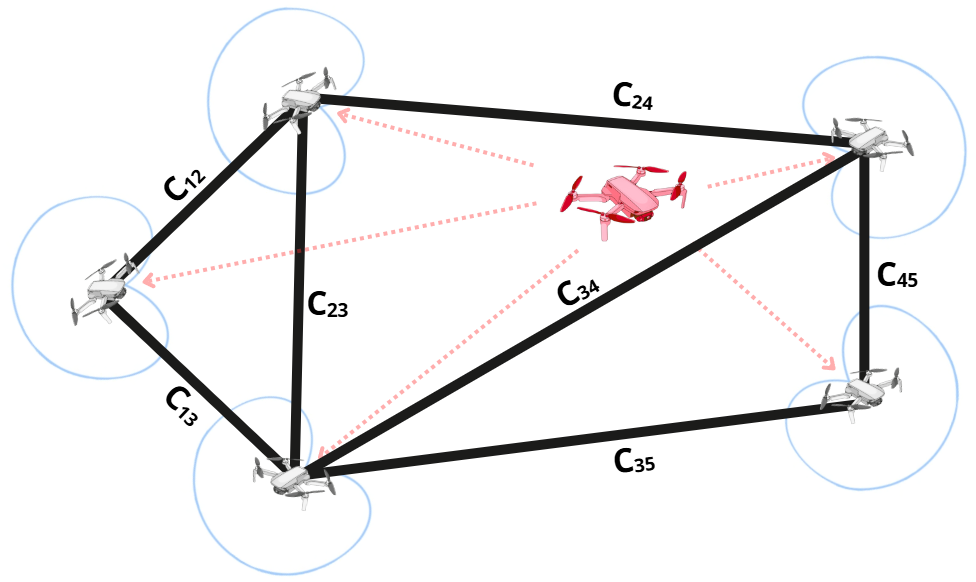}
    \caption{Representation of UAV swarm as a communication graph.}
    \label{fig:graph}
\end{figure}

Each UAV \(i\) has position \(p_i = (x_i, y_i)\) and antenna orientation \(\theta_i \in [0^\circ, 360^\circ]\). UAVs move with constant velocity during each epoch, avoiding collisions through altitude separation. Communication links between UAVs are represented as a graph (Fig.~\ref{fig:graph}), where nodes represent UAVs and edges correspond to link capacities \(C_{ij}\).

The communication model integrates path loss, received power, and link capacity. Path loss is expressed as:
\begin{equation}
L = L_{d_0} + 10n \log_{10}\left(\frac{d}{d_0}\right) + \chi,
\end{equation}
and the received power as:
\begin{equation}
P_{Rx} = P_{Tx} + G_{Tx} + G_{Rx} - L.
\end{equation}
The link capacity follows the Shannon-Hartley theorem:
\begin{equation}
C = B \log_2(1 + \text{SINR}),
\quad
\text{SINR} = \frac{P_{Rx}}{P_{\text{int}} + P_{\text{noise}}}.
\end{equation}

The objective function for each timeslot is defined as:
\begin{equation}
F(p, \theta) = (C_{\text{avg}})^\alpha \cdot (C_{\text{min}})^\beta,
\end{equation}
where \(C_{\text{avg}}\) and \(C_{\text{min}}\) denote the average and minimum link capacities, and \(\alpha,\beta\) are weighting coefficients. The global objective over an epoch is the mean of all timeslots:
\begin{equation}
\max F_{\text{objective}} = \frac{1}{T} \sum_{t=1}^{T} 
[(C_{\text{avg}, t})^\alpha (C_{\text{min}, t})^\beta].
\end{equation}

\subsection{Proposed Algorithmic Framework}

The optimization seeks the vector 
\(\mathbf{v}^* = [p_{it}, \theta_{it}]_{i=1..N, t=0..T}\)
that maximizes \(F_{\text{objective}}\). Three independent approaches - GA, SL, and RL - are employed to identify \(\mathbf{v}^*\).

\subsubsection{Genetic Algorithm (GA)}

The GA serves as the baseline optimization method, offering a balance between solution quality and computational cost. Each chromosome encodes UAV positions and antenna orientations. Fitness is evaluated using \(F_{\text{objective}}\), and standard GA operators (selection, crossover, mutation) guide the search for the best configuration. The GA also generates datasets for training SL models, providing a bridge toward learning-based approaches.

\subsubsection{Supervised Learning (SL)}

Supervised models, including K-Nearest Neighbors (KNN), Random Forest (RF), and XGBoost, are trained using datasets produced by the GA. Input features include jammer coordinates and initial UAV positions, while outputs represent optimized final UAV positions. Models are trained on discrete data but evaluated on continuous configurations, enabling generalization to unseen spatial patterns.

\subsubsection{Reinforcement Learning (RL)}

RL introduces adaptability to dynamic environments. A custom OpenAI Gym environment models the UAV-jammer interaction, where the swarm learns to maximize communication through trial and error. The problem is formulated as a Markov Decision Process (MDP) with:
\begin{itemize}
    \item \textbf{State:} UAV positions, orientations, jammer position, and link capacities;
    \item \textbf{Action:} Adjustments in positions and antenna angles;
    \item \textbf{Reward:} Based on \(F_{\text{objective}}\), encouraging high capacity and minimal interference.
\end{itemize}
The PPO algorithm is employed for policy optimization, initialized via supervised pretraining to accelerate convergence.

\subsection{Movement Scenarios and Collision Rules}

Three movement scenarios are evaluated:  
(a) static UAVs optimizing only antenna directions;  
(b) static formation with positional adjustment; and  
(c) dynamic swarm progression over time.  
Additionally, three collision-handling rules are applied:  
(i) collisions allowed: used as a baseline without spatial constraints;  
(ii) final position collision-free: allowing temporary overlaps during motion but enforcing separation at the end of each epoch; and  
(iii) trajectory collision-free: prohibiting collisions at any time to ensure fully realistic swarm movement.  
These variations allow consistent comparison across models and highlight robustness under diverse operational constraints.

\subsection{Evaluation Metrics}

Performance is evaluated through three main merit metrics: minimum link capacity ($C_{\text{min}}$), average link capacity ($C_{\text{avg}}$), and the objective function fitness. Complementary metrics include computational time, Mean Euclidean Distance (MED), and collision rate per epoch. These collectively assess both the communication efficiency and physical feasibility of the optimized swarm behavior. In most cases, these metrics are used in their averaged form, represented as $\overline{C_{\text{min}}}$, $\overline{C_{\text{avg}}}$, and $\overline{\text{Fitness}}$, computed over all runs or samples within the dataset.

\section{Results }

This section summarizes the main experiments conducted throughout the study and discusses the corresponding outcomes. Each stage, from the Genetic Algorithm (GA) to Supervised Learning (SL) and Reinforcement Learning (RL). The detailed quantitative results and final performance comparison across all algorithms are presented in Subsection~\ref{subsec:acomparison}.

Table~\ref{tab:uav_optimization_parameters} summarizes the main parameters applied throughout the UAV swarm optimization experiments. These are grouped into three categories: UAV/jammer system, propagation model, and spatial–temporal parameters, which together define both the communication conditions and the physical simulation environment illustrated in Figure~\ref{fig:jammingcomm}.

\begin{table}[H]
\centering
\caption{Parameters applied in the UAV swarm optimization experiments.}
\label{tab:uav_optimization_parameters}
\renewcommand{\arraystretch}{1.2}
\begin{tabular}{lc}
\hline
\textbf{Parameter} & \textbf{Value} \\ \hline
\multicolumn{2}{c}{\textbf{UAV / Jammer System Parameters}} \\ \hline
Wavelength $\lambda_0$ & 0.125 m \\
Bandwidth & $20 \times 10^{6}$ Hz \\
UAV transmission power & 20 dBm \\
Jammer interference power & 100 dBm \\
Noise power & -100 dBm \\
Minimum distance between UAVs & 20 m \\ 
Number of UAVs & 4 \\ \hline
\multicolumn{2}{c}{\textbf{Propagation Model Parameters}} \\ \hline
Reference distance $d_0$ & 1 m \\
Path loss at reference distance $L_{d_0}$ & 30 dB \\
Path loss exponent $n$ & 2.7 \\ \hline
\multicolumn{2}{c}{\textbf{Spatial and Temporal Parameters}} \\ \hline
Timeslot cell size & 60 m × 60 m \\
Epoch spatial length & 120 m \\ 
Maximum UAV speed $v_{\max}$ & 20 m/s \\
Epoch duration $T_{\text{epoch}}$ & 13.4 s \\
Number of timeslots per epoch & 6 \\
Timeslot duration $T_{\text{ts}}$ & 2.23 s \\
Grid resolution & 4 × 4 \\ 
\hline
\end{tabular}
\end{table}

\subsection{GA Results}

\subsubsection{Static Scenario with Antenna Directionality}

\begin{figure}[htbp]
    \centering
    \includegraphics[width=0.35\textwidth]{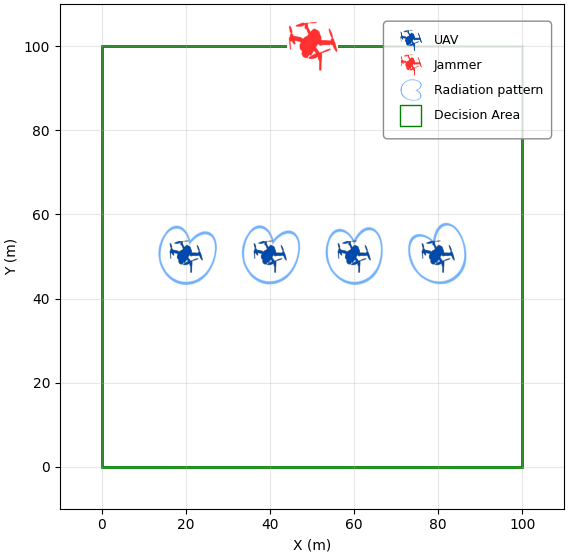}
    \caption{Formation 1.}
    \label{fig:static_scenario}
\end{figure}

This experiment validates the functionality of the Genetic Algorithm (GA) in optimizing UAV antenna orientation under jamming conditions. Four swarm formations were tested with random initial orientations, one of which is illustrated in Figure~\ref{fig:static_scenario}. The baseline corresponds to a simple alignment strategy where each UAV directs its antenna null straight toward the jammer. Compared to this baseline, the GA consistently achieved superior configurations, with fitness improvements between +112\% and +210,000\% and average capacity gains up to +135,723\%. These results confirm the correctness of the implementation and demonstrate that the GA effectively enhances communication performance even in static configurations, establishing a solid foundation for subsequent dynamic scenarios.

\subsubsection{Static Scenario with Exploitation of Antenna Directionality and Positions}

In this scenario, UAVs were allowed to adjust both their positions and antenna orientations within a fixed area to improve network topology and communication efficiency. Two optimization strategies were tested: one performing full optimization of both parameters, and another restricting optimization to positions only, with antennas fixed toward the jammer through null-steering.

\begin{figure}[htbp]
    \centering
    \includegraphics[width=0.35\textwidth]{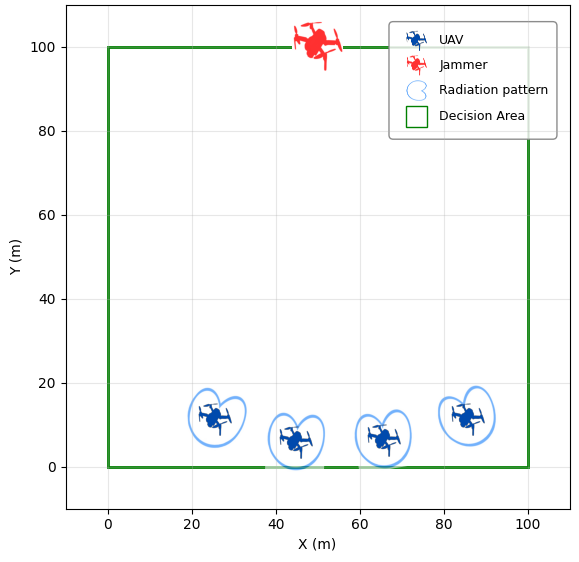}
    \caption{Optimized UAV formations for the position-only strategy under Pop. size = 400, Generations = 400.}
    \label{fig:output3}
\end{figure}

Multiple combinations of Genetic Algorithm (GA) parameters were evaluated, varying the population size and number of generations to assess their effect on convergence and performance. Figure~\ref{fig:output3} illustrates the optimized UAV formation for the position-only strategy with a population size of 400 and 400 generations, which yielded the best overall performance.

Results showed that the position-only strategy converged faster and achieved comparable or superior fitness values (up to 9.37$\times$10$^{6}$) with significantly lower computational cost. Since antenna orientation is already optimally determined by the jammer’s direction, only UAV positions were optimized in subsequent experiments. This simplification reduces dimensionality, improves convergence speed, and enhances the model’s suitability for real-time swarm control.

\subsubsection{Scenario with a Progressing Swarm}
\label{subsec:progressing_swarm}

This scenario represents the transition from static configurations to a dynamic and mission-oriented swarm behavior. Using the Genetic Algorithm (GA), UAVs progressively move toward a target area while maintaining communication efficiency under jamming conditions.  

\begin{figure}[h!]
    \centering
    \includegraphics[width=0.49\textwidth]{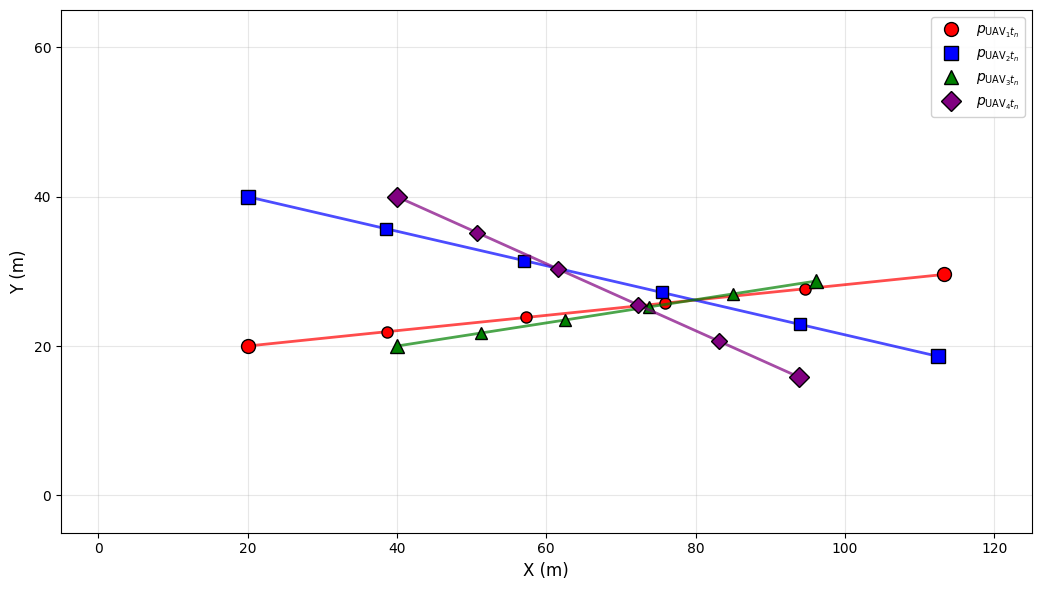}
    \caption{UAV swarm trajectories under Collision Rule 1 - all collisions permitted.}
    \label{fig:single_epoch}
\end{figure}

In the initial single-epoch setup, the objective was to validate whether the GA could coordinate UAV trajectories within one mission phase. Each UAV adjusted its position over six timeslots to maximize the global fitness function. This process was repeated for all three collision-handling rules to assess the swarm’s adaptability under different spatial constraints. Figure~\ref{fig:single_epoch} illustrates an example corresponding to \textit{Collision Rule~1}, where all collisions are permitted, showing how the swarm advances coherently within the operational area while respecting the jammer’s interference constraints.

\begin{figure}[h!]
    \centering
    \includegraphics[width=0.47\textwidth]{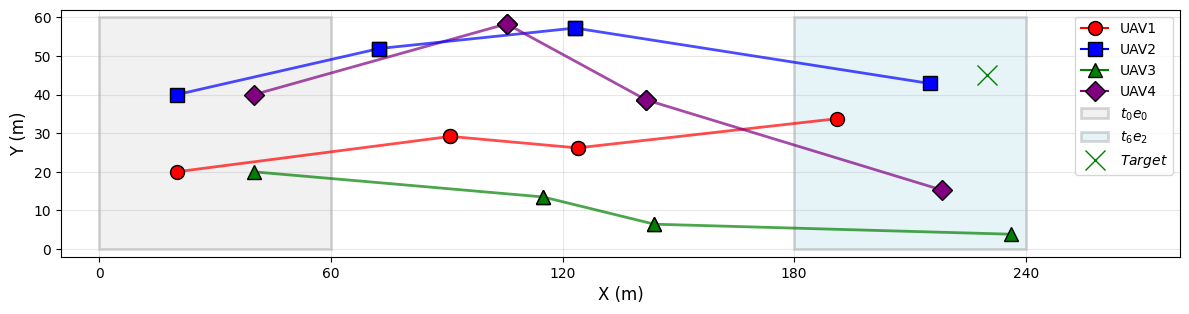}
    \caption{Example of a progressing swarm over multiple epochs.}
    \label{fig:multi_epoch}
\end{figure}

After confirming stable performance within one epoch, the framework was extended to a multi-epoch version that enables continuous swarm progression toward a designated target. In this configuration, the final positions from one epoch automatically define the initial conditions of the next, allowing seamless advancement over multiple mission stages. The process continues until the swarm reaches the target region, as illustrated in Figure~\ref{fig:multi_epoch}.  

The main parameters used in this implementation, including population size, number of generations, and geometric limits for the UAVs and jammer, are summarized in Table~\ref{tab:uav_params_multi_epoch2}. The \textit{Best Solution Matrix} stores the optimal UAV positions obtained for each timeslot and epoch, representing the full trajectory plan of the swarm, while the \textit{Fitness Vector} records the corresponding fitness values achieved in each epoch, providing a quantitative measure of the optimization performance over time.

\begin{table}[h!]
\centering
\caption{GA input and output parameters for the multi-epoch UAV swarm optimization.}
\label{tab:uav_params_multi_epoch2}
\renewcommand{\arraystretch}{1.1}
\begin{tabular}{lc}
\hline
\textbf{Parameters} & \textbf{Value} \\ \hline
\multicolumn{2}{c}{\textbf{Input Parameters}} \\ \hline
Number of Timeslots & 6 \\
Initial UAV Positions & $x,y \in [0, 60]$\,m \\
Target Position & $y \in [60, 120]$\,m \\
Jammer Position & $(x, 500)$\,m \\
Population Size & 50 \\
Generations & 50 \\ \hline
\multicolumn{2}{c}{\textbf{Output Parameters}} \\ \hline
Best Solution Matrix &  \\
Fitness Vector &  \\
Number of Epochs & \\ \hline
\end{tabular}
\end{table}

Both single- and multi-epoch implementations confirm that the GA effectively coordinates swarm movement under interference, generating smooth and feasible trajectories toward the target. The multi-epoch version, in particular, introduces a scalable structure capable of adapting to larger missions without additional manual configuration. The quantitative evaluation of GA performance in these scenarios is discussed later in the algorithm comparison section.

\subsubsection{Adaptive Movement Model}
\label{subsec:amovcomparison}

This subsection introduces the Adaptive Movement Model, a rotation-based mechanism that extends the Genetic Algorithm (GA) to handle UAV movement toward arbitrary target positions in any direction, overcoming the corridor constraint of previous scenarios. By rotating the entire environment around the swarm’s center of mass, the problem is transformed into an equivalent configuration aligned with the optimization corridor. Once the optimal trajectories are computed, the resulting \textit{Best Solution Matrix} is rotated back to the original coordinate system.

\begin{figure}[h!]
    \centering
    \includegraphics[width=0.45\textwidth]{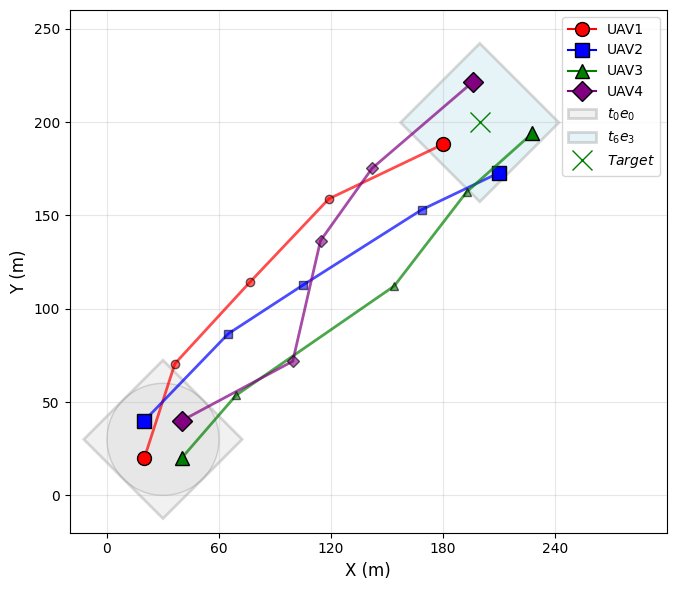}
    \caption{Example of a progressing swarm scenario under the Adaptive Movement Model.}
    \label{fig:caminho1222}
\end{figure}

As shown in Figure~\ref{fig:caminho1222}, UAVs are initially positioned within a circle of radius 30\,m inscribed in the first timeslot square. This constraint ensures that all UAVs remain within valid spatial boundaries during rotation, maintaining feasible trajectories.  

The corresponding input and output parameters are summarized in Table~\ref{tab:uav_params}, which are mostly identical to those of the previous multi-epoch scenario, with the addition of the \textit{Rotation Angle}~($\theta$), representing the applied coordinate transformation.

\begin{table}[h!]
\centering
\caption{Input and output parameters used in the Adaptive Movement Model.}
\label{tab:uav_params}
\renewcommand{\arraystretch}{1.1}
\begin{tabular}{lc}
\hline
\textbf{Parameters} & \textbf{Value} \\ \hline
\multicolumn{2}{c}{\textbf{Input Parameters}} \\ \hline
Number of Timeslots & 6 \\
Initial UAV Positions & $(x-30)^2+(y-30)^2 \le 30^2\ $ \\
Target Position & $(x,y)$ \\
Jammer Position & $(x,y)$ \\
Population Size & 50 \\
Generations & 50 \\ \hline
\multicolumn{2}{c}{\textbf{Output Parameters}} \\ \hline
Best Solution Matrix &  \\
Fitness Vector &  \\
Number of Epochs &  \\
Rotation Angle &  \\ \hline
\end{tabular}
\end{table}

This model ensures that by effectively optimizing a single epoch, the swarm’s overall trajectory can be reliably extended across multiple epochs with consistent performance. In other words, the Adaptive Movement Model confirms that the optimization principles validated in one epoch remain applicable throughout the mission, maintaining stable behavior as the swarm advances toward its target.

\subsection{Supervised Learning (SL)}
\label{subsec:slb}

In this section, three GA-generated datasets were used to train supervised learning (SL) models, each corresponding to a distinct collision-handling rule. The training datasets were discrete, containing over 305,760 samples each, while three continuous GA-generated test sets of 2,400 unseen samples were used to determine whether SL models can generalize beyond the discrete, collision-free training data and accurately predict optimal UAV swarm configurations in unseen continuous environments.

\subsubsection{Interpolation}

The K-Nearest Neighbors (KNN) algorithm was applied as an interpolation-based SL method to estimate optimal UAV positions from the reference data. KNN performance was analyzed under different values of \(k\) and collision-handling policies, and results were compared with the GA and a random baseline.

For \textit{Collision Rule 1 (All Collisions Permitted)}, KNN achieved the highest performance at \(k=2\), surpassing both GA and random baselines across all merit metrics, indicating that discrete interpolation can effectively approximate optimal solutions.

Under \textit{Collision Rule 2 (Final Position Collision-Free)}, KNN outperformed the random approach but did not reach GA’s performance levels, while collision rates exceeded 50\%.

For \textit{Collision Rule 3 (Trajectory Collision-Free)}, KNN performance degraded sharply as \(k\) increased, with fitness values approaching zero and collisions reaching 100\%.

Overall, results demonstrate that while KNN interpolation can efficiently reproduce GA-like behavior in unconstrained conditions, its ability to handle collision constraints and dynamic spatial relationships is limited—validating the need for more adaptive learning approaches in subsequent reinforcement learning experiments.

\subsubsection{Regression}

Regression-based supervised learning was applied to predict UAV swarm configurations using the datasets generated by the GA. Among the tested models, Random Forest (RF) achieved the most consistent and superior results across all metrics and scenarios, clearly outperforming other regression methods such as XGBoost.

For \textit{Collision Rule 1 (All Collisions Permitted)}, RF achieved exceptionally high fitness and communication capacities, outperforming both the GA and the random baseline by several orders of magnitude. This demonstrates RF’s strong capacity to learn and generalize the patterns encoded in the GA-generated dataset when no spatial restrictions are imposed.

Under \textit{Collision Rule 2 (Final Position Collision-Free)}, RF maintained competitive performance in terms of communication metrics but exhibited collision rates above 50\%, indicating that, while the model could replicate efficient positioning, it lacked spatial awareness to enforce separation constraints. 

In the most constrained setup, \textit{Collision Rule 3 (Trajectory Collision-Free)}, RF performance degraded substantially, with fitness and capacity values dropping close to the random baseline. The model failed to adapt to the strict spatial limitations, confirming that supervised learning alone is insufficient for collision-aware optimization.

In summary, Random Forest demonstrated strong generalization and high predictive accuracy in unconstrained scenarios but struggled to maintain feasible solutions as spatial restrictions increased. These results confirm the need for adaptive models, such as reinforcement learning, to integrate collision awareness and continuous decision-making beyond static pattern replication.

\subsection{RL Results}

The RL experiments used the same GA-generated datasets from the SL stage for pretraining, enabling faster convergence.

\begin{figure}[h!]
    \centering
    \includegraphics[width=0.40\textwidth]{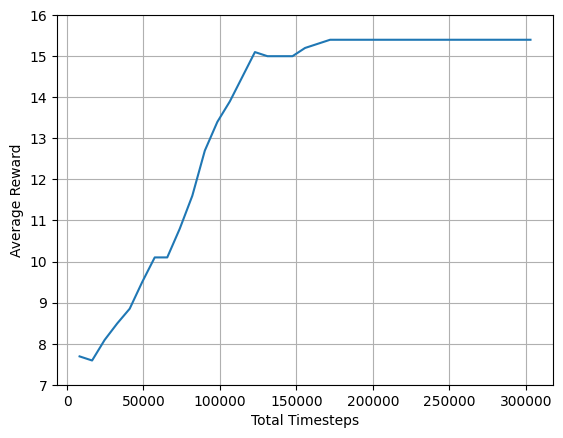}
    \caption{Reward Evolution under Fixed Positions (Rule 1).}
    \label{fig:rl1}
\end{figure}

\begin{figure}[h!]
    \centering
    \includegraphics[width=0.40\textwidth]{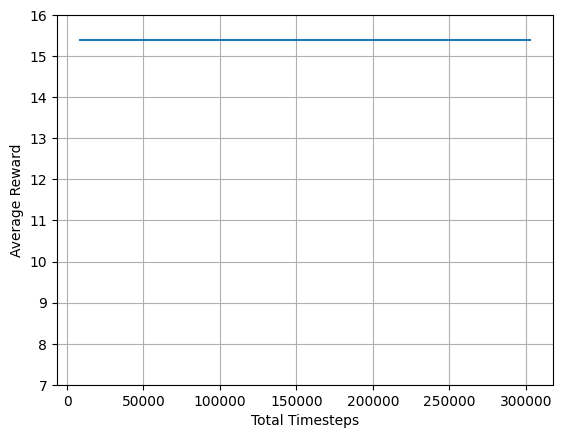}
    \caption{Reward Progress with Generalized Training (Rule 1).}
    \label{fig:rl11}
\end{figure}

As illustrated in Figures~\ref{fig:rl1} and~\ref{fig:rl11}, training followed a two-phase process: first under fixed UAV and jammer positions, and then with randomized initial and target conditions to test generalization.  

Under \textit{Collision Rule 1}, where collisions were fully permitted, the RL agent achieved the highest overall performance, matching or surpassing the GA in fitness and link capacities. As seen in Figures~\ref{fig:rl1} and~\ref{fig:rl11}, the reward steadily increases during the fixed-position phase until convergence, confirming stable policy learning. When generalized to randomized conditions, the reward remains high and consistent, demonstrating successful transfer learning and robust generalization to new scenarios.  

\begin{figure}[h!]
    \centering
    \includegraphics[width=0.40\textwidth]{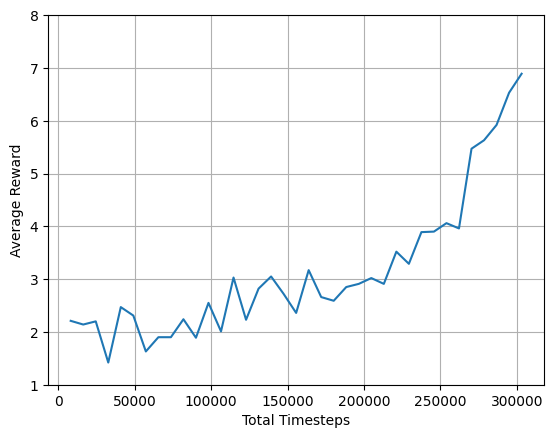}
    \caption{Reward Evolution under Fixed Positions (Rule 2).}
    \label{fig:rl2}
\end{figure}

\begin{figure}[h!]
    \centering
    \includegraphics[width=0.40\textwidth]{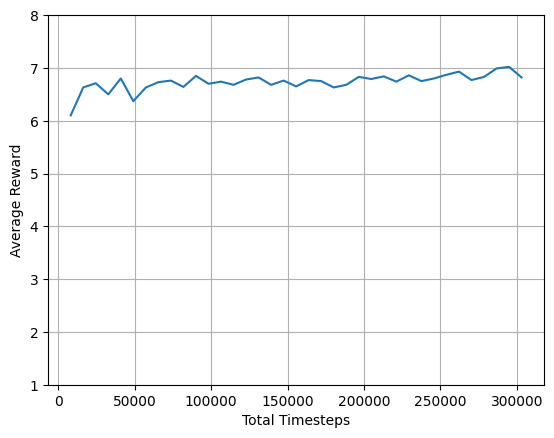}
    \caption{Reward Progress with Generalized Training (Rule 2).}
    \label{fig:rl22}
\end{figure}

When \textit{Collision Rule 2} was introduced, enforcing collision-free final positions, the agent successfully maintained spatial safety with zero collisions while preserving stable communication levels, demonstrating effective policy generalization. As shown in Figures~\ref{fig:rl2} and~\ref{fig:rl22}, the reward increased steadily under fixed positions and later stabilized after a brief drop during generalization, showing that the agent adapted effectively to new spatial conditions while maintaining collision-free behavior.  

\begin{figure}[h!]
    \centering
    \includegraphics[width=0.40\textwidth]{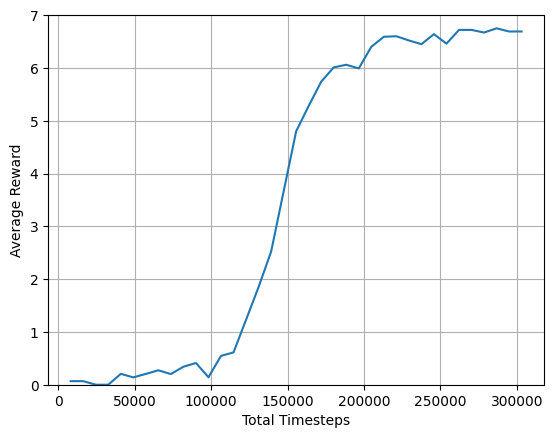}
    \caption{Reward Evolution under Fixed Positions (Rule 3).}
    \label{fig:rl3}
\end{figure}

\begin{figure}[h!]
    \centering
    \includegraphics[width=0.40\textwidth]{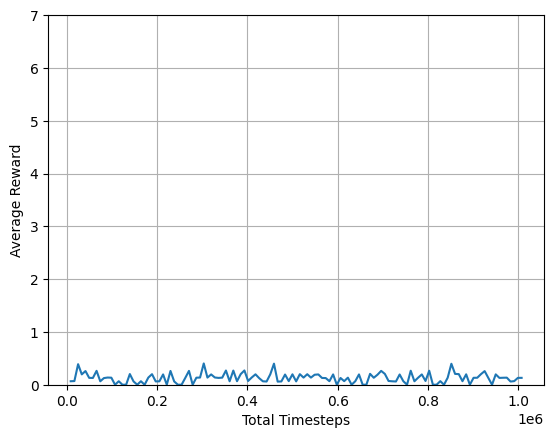}
    \caption{Reward Progress with Generalized Training (Rule 3).}
    \label{fig:rl33}
\end{figure}

In the most restrictive case, \textit{Collision Rule 3}, where collisions were prohibited throughout the entire trajectory, performance decreased due to stricter constraints, but the agent continued to produce feasible, collision-free trajectories. As illustrated in Figures~\ref{fig:rl3} and~\ref{fig:rl33}, the reward increased under fixed positions but became unstable and lower during generalization, reflecting the difficulty of maintaining safe, collision-free trajectories under such conditions.  

Overall, RL provided a strong balance between performance and feasibility, showing that an adaptive learning-based approach can approach the optimization quality of the GA while operating autonomously and with far lower computational cost. A full comparison of all algorithms is presented in the following section.

\subsection{Algorithm Comparison}
\label{subsec:acomparison}

This section compares all implemented algorithms under the three defined collision rules, assessing their performance, stability, and adaptability to spatial constraints, summarizing the key findings from the previous result subsections.

\begin{table}[H]
\centering
\caption{Comparison of algorithms — Collision Rule 1.}
\label{tab:algorithm_comparison1}
\renewcommand{\arraystretch}{1.1}
\begin{tabular}{lrrr}
\hline
\textbf{Algorithm} & $\mathbf{\overline{\text{Fitness}}}$ & $\mathbf{\overline{C_{\text{avg}}}\text{ (bps)}}$ & $\mathbf{\overline{C_{\min}}\text{ (bps)}}$ \\
\hline
KNN (K=2) & 8.14e15 & 2.94e7 & 2.64e7 \\
RF & 7.33e14 & 6.29e6 & 3.03e6 \\
RL & 2.70e15 & 2.12e7 & 2.11e7 \\
GA (Ref) & 3.29e8 & 1.81e4 & 7.42e3 \\
Random & 1.51e7 & 6.63e3 & 1.83e3 \\
\hline
\end{tabular}
\end{table}

Under Rule~1, where collisions were fully permitted, all algorithms achieved high fitness due to the lack of spatial constraints.  
KNN obtained the highest values by interpolating the discrete GA-generated data, while RL approached these results and even surpassed the continuous GA baseline.  
This confirms that RL can generalize from discrete training sets to continuous environments despite the inherent differences in data representation.

\begin{table}[H]
\centering
\caption{Comparison of algorithms — Collision Rule 2.}
\label{tab:algorithm_comparison2}
\renewcommand{\arraystretch}{1.1}
\begin{tabular}{lrrrr}
\hline
\textbf{Algorithm} & $\mathbf{\overline{\text{Fitness}}}$ & $\mathbf{\overline{C_{\text{avg}}}\text{ (bps)}}$ & $\mathbf{\overline{C_{\min}}\text{ (bps)}}$ & \textbf{Collision (\%)} \\
\hline
KNN (K=1) & 3.59e7 & 8.10e3 & 3.09e3 & 25.95 \\
RF & 2.29e7 & 4.22e3 & 1.58e3 & 51.75 \\
RL & 1.13e7 & 5.30e3 & 1.79e3 & 0.00 \\
GA (Ref) & 2.25e8 & 1.62e4 & 5.89e3 & 0.00 \\
Random & 1.73e6 & 8.63e2 & 2.41e2 & 85.58 \\
\hline
\end{tabular}
\end{table}

Under Rule~2, where only final collisions were restricted, the supervised models (KNN and RF) attempted to reproduce GA-generated trajectories, leading to collision rates of 26\% and 52\%, respectively.  
RL maintained zero collisions and valid trajectories, achieving roughly 30\% of the GA’s fitness but with significantly lower computational time once trained.  
This illustrates the trade-off between GA’s exhaustive optimization and RL’s faster, autonomous decision-making.

\begin{table}[H]
\centering
\caption{Comparison of algorithms — Collision Rule 3.}
\label{tab:algorithm_comparison3}
\renewcommand{\arraystretch}{1.1}
\begin{tabular}{lrrrr}
\hline
\textbf{Algorithm} & $\mathbf{\overline{\text{Fitness}}}$ & $\mathbf{\overline{C_{\text{avg}}}\text{ (bps)}}$ & $\mathbf{\overline{C_{\min}}\text{ (bps)}}$ & \textbf{Collision (\%)} \\
\hline
KNN (K=1) & 1.06e7 & 3.21e3 & 1.32e3 & 25.47 \\
RF & 1.86e6 & 4.17e2 & 2.21e2 & 94.54 \\
RL & 7.05e6 & 4.89e3 & 1.21e3 & 0.00 \\
GA (Ref) & 1.76e7 & 5.90e3 & 2.64e3 & 0.00 \\
Random & 1.73e6 & 8.63e2 & 2.41e2 & 85.58 \\
\hline
\end{tabular}
\end{table}

Finally, under Rule~3 — prohibiting collisions throughout the trajectories — all deterministic methods showed reduced performance.  
The RL agent maintained zero collisions and stable capacities, achieving results moderately below the GA baseline.  
These findings reinforce the consistent trade-off between raw optimization performance (GA) and operational efficiency (RL), confirming that both methods complement each other under different mission constraints.

\section{Conclusion and Future Work}

\subsection{Conclusion}

With the continuous expansion of UAV applications in communication, surveillance, and defense, ensuring reliable and resilient swarm operation under jamming conditions has become a critical research challenge.

This work addressed the challenge of answering the question: 
\textit{``Can a swarm of UAVs effectively overcome jamming while maintaining communication and achieving mission success?''}

Through the integration of Genetic Algorithm (GA) optimization, Supervised Learning (SL), and Reinforcement Learning (RL) within a unified framework, the study demonstrated that such capability is not only feasible but also scalable, not only in static and simplified scenarios but also in dynamic environments where UAV path planning, antenna orientation behaviors, and swarm formation are progressively integrated to preserve communication efficiency under jamming conditions.

The proposed unified framework, developed in this work, introduces a structured mission model based on a system of epochs and timeslots, in which the complete mission path is divided into multiple sequential epochs. Within each epoch, the UAVs’ performance, including their positions, trajectories, and antenna orientations, is evaluated at every timeslot, allowing continuous assessment and adaptation of communication efficiency and interference mitigation. This progressive design supports the stepwise inclusion of path planning, antenna orientation, and swarm formation, enabling a fair and systematic comparison of all implemented algorithms under identical yet increasingly complex mission conditions.

The results showed that the GA effectively explored large solution spaces, generating diverse and high-quality datasets for training subsequent learning models. As scenario complexity increased, particularly with the introduction of collision rules, the GA remained stable and reliable, producing collision-free trajectories with the chosen parameters. However, its high computational cost and static nature limited its adaptability to dynamic conditions, motivating the use of learning-based methods for faster decision-making.

SL models, trained on discrete datasets generated by the GA, successfully reproduced known configurations and achieved high performance in collision-free scenarios. They effectively transferred the knowledge learned from discrete GA solutions into continuous environments. However, when collision constraints and dynamic interference were introduced, these models frequently failed to adapt, leading to unstable trajectories and mission failure. This highlights the limited adaptability of purely data-driven learning in unpredictable swarm conditions.

The RL approach, initialized with GA-generated datasets and trained using Proximal Policy Optimization (PPO), demonstrated strong adaptability through online interaction with the environment. It successfully learned to complete missions without collisions, maintaining feasible trajectories and consistent communication under dynamic conditions. Although its overall fitness values were lower than the GA’s, especially as collision complexity increased, the RL agent achieved real-time adaptability and reduced computational overhead. This reflects a clear trade-off between optimization performance and operational efficiency, emphasizing RL’s potential for autonomous, time-constrained swarm missions.

Furthermore, the introduction of the Adaptive Movement Model demonstrated that the same optimization structure can generalize to arbitrary movement directions. By rotating the scenario around the swarm’s center of mass and validating results across epochs, it was shown that a single well-performing epoch can be extended to any trajectory or orientation, confirming the scalability and spatial generalization of the proposed approach.

Overall, the findings confirm that UAV swarms equipped with \textbf{null-steering antennas} and guided by intelligent optimization algorithms can effectively mitigate jamming while maintaining stable communication and formation cohesion. While certain algorithms, such as GA, perform best in static or fully known environments, others like RL excel in dynamic and unpredictable conditions, achieving real-time adaptability. The proposed framework not only unifies multiple optimization paradigms but also provides a transparent and reproducible foundation for future anti-jamming swarm research, publicly available through the associated GitHub repository.

\subsection{Future Work}

Building on the promising results obtained in this research, several directions can be explored to extend and enrich the proposed framework. A natural next step involves expanding the optimization analysis to include comparisons with additional metaheuristic algorithms beyond GA, as well as exploring alternative Supervised and Reinforcement Learning paradigms beyond those studied in this work, to better understand the trade-offs between convergence speed, computational cost, and communication efficiency in complex swarm environments.

Another relevant direction is the extension of the current two-dimensional model into a fully three-dimensional space, where UAVs operate with altitude variations and under more realistic aerodynamic and propagation constraints. This enhancement would allow the framework to capture the spatial dynamics of real-world missions, particularly in environments with obstacles or varying terrain elevation.

The antenna model could also be further diversified by testing alternative configurations beyond null-steering, such as adaptive beamforming. Comparing the performance of these antenna systems under identical conditions would help identify the most efficient configuration for resilience against jamming.

Additionally, introducing more dynamic and uncertain adversarial behavior would bring the simulation closer to real operational conditions. Future scenarios may include a jammer that gradually changes position or power level during the mission, without the UAVs having prior knowledge of its trajectory. This would test the ability of learning-based methods, particularly reinforcement learning, to adapt in real time to evolving interference patterns.

Finally, the framework could be extended to a multi-agent setting, where multiple UAVs learn cooperatively under decentralized control. This would allow each UAV to make decisions based on local observations while contributing to a shared communication objective, enabling the study of cooperative behaviors, coordination strategies, and swarm-level adaptability under realistic and dynamic mission conditions.

Overall, this field remains in active development, offering numerous opportunities for future research and technological advancement.

\end{document}